\documentclass[journal]{IEEEtran}

\usepackage{times}
\usepackage{epsfig}
\usepackage{graphicx}
\usepackage{amsmath}
\usepackage{amssymb}

\usepackage{subfigure}

\usepackage{algorithm}  
\usepackage{algorithmicx}  
\usepackage{algpseudocode}  
\usepackage{booktabs}
\usepackage{multirow}
\usepackage{color}
\usepackage{gensymb}
\usepackage{hyperref}
\ifCLASSINFOpdf
\else
\fi
\begin{document}
\title{RiWNet: A moving object instance segmentation Network being Robust in adverse Weather conditions}

\author{Chenjie Wang, Chengyuan Li, Bin Luo, Wei Wang, Jun Liu
	\thanks{C. Wang, C. Li, B. Luo, W. Wang, J. Liu are with State Key Laboratory of Information Engineering in Surveying, Mapping and Remote Sensing, Wuhan University, Wuhan 430072, China (e-mail:
	{\{wangchenjie, lichengyuan, luob, kinggreat24, liujunand\}@whu.edu.cn)}
}%
}% <-this % stops a space

\maketitle

\begin{abstract}
Segmenting each moving object instance in a scene is essential for many applications. But like many other computer vision tasks, this task performs well in optimal weather, but then adverse weather tends to fail. To be robust in weather conditions, the usual way is to train network in data of given weather pattern or to fuse multiple sensors. We focus on a new possibility, that is, to improve its resilience to weather interference through the network's structural design. First, we propose a novel FPN structure called RiWFPN with a progressive top-down interaction and attention refinement module. RiWFPN can directly replace other FPN structures to improve the robustness of the network in non-optimal weather conditions. Then we extend SOLOV2 to capture temporal information in video to learn motion information, and propose a moving object instance segmentation network with RiWFPN called RiWNet. Finally, in order to verify the effect of moving instance segmentation in different weather disturbances, we propose a VKTTI-moving dataset which is a moving instance segmentation dataset based on the VKTTI dataset, taking into account different weather scenes such as rain, fog, sunset, morning as well as overcast. The experiment proves how RiWFPN improves the network's resilience to adverse weather effects compared to other FPN structures. We compare RiWNet to several other state-of-the-art methods in some challenging datasets, and RiWNet shows better performance especially under adverse weather conditions.

%Then based on RiWFPN and SOLOV2, a moving object instance segmentation network called RiWNet is proposed, which uses a ConvLSTM-based structural design to learn motion information. 

\end{abstract}


% \fi
\IEEEpeerreviewmaketitle

\section{Introduction}
Detecting and segmenting out every moving object instance in the dynamic scene is key to safe and reliable autonomous driving. It also supports the task of dynamic visual SLAM~\cite{MaskFusion2018, DymSLAM2021}, dynamic object obstacle avoidance~\cite{dynamicobstacles2008}, video surveillance~\cite{AnAdvanced2014} and decision-making of autonomous driving~\cite{Motionplanning2008}. In recent years, there are many deep learning methods~\cite{xie2019object,dave2019towards,muthu2020motion} that can segment each moving object instance well in the dynamic scene. However, like many perception applications including the semantic segmentation~\cite{RobustSemantic2019} and object detection~\cite{mirza2021robustness} of the camera streams, these moving instance segmentation methods performs well in good weather conditions and are likely to fail in non-optimal weather conditions. In the real environment, changing weather conditions often appear unexpectedly, which is one of the more challenging problems to mitigate against in perception systems~\cite{fursa2021worsening}. For example, in fog, snow, rain, at night or even in blinding sunlight, camera images are disturbed by adverse weather effects, causing the perception performance to decreases enormously. 

To obtain a robust perception effect in degrading weather scenarios such as rain, fog and night, there are several ways proposed in recent years. Some methods~\cite{RainRemoval2019,Semantic2019} simulate the impact of varying weather pattern and use these data for network training. However, it is almost impossible to simulate the impact of all weather pattern in the training data because of complex and changing weather conditions. These methods also increase a large amount of training data and therefore lead to an increased training burden. There are some methods ~\cite{AutomaticAdaptation2019,UnsupervisedMonocular2020} that use domain adaptation methods to adapt methods that perform well in the source domain (good weather) to the target domain (different weather scenarios), and still perform well in different weather domains. Such methods increase the difficulty of training the network and is possible to make it difficult for the network to converge. Considering that different sensor types perform differently in different weather scenarios, some methods~\cite{RobustSemantic2019,SeeingThrough2020} fuse data of diverse sensors to obtain more reliable results. Limited by factors such as cost and equipment limitations, in most cases it is difficult to have enough types of sensors to be used at the same time. The video-segmentation based method~\cite{pfeuffer2020robust} capture temporal information of previous frames to compensate current segmentation errors. However, the method of video processing inevitably come with a large amount of computational burden and memory cost which greatly increases the inference time, making it difficult to run in real time. 

In this paper, we focus on another possibility, improve the robustness of the network to different weather effects through the design of the network structure. This method does not need to add additional structure to the network. Meanwhile, the training input does not need to limit a single weather interference pattern, that is, to train images of different weather patterns together, and get very robust results in each weather condition. First, we propose a novel FPN module called RiWFPN (Robust in Weather conditions Feature Pyramid Network), which composed of a progressive top-down interaction module and attention refinement module. Just using RiWFPN to replace the existing FPN structure can make the network more robust against diverse weather disturbances. The idea is that the structure information of the object in the image can represent the object well. Even if the main body of the object has been disturbed by the weather effect, we think that trying to preserve and strengthen the structural information in the image can help discover the object. RiWFPN uses a progressive top-down interaction module to make feature maps from different scales of pyramid structure “cleaner” and introduce a lot of semantic and spatial information. And then it uses attention refinement module to refine the abundant information in each layer and enhance low-frequency components of network to refine the structure information, to make moving objects easier to discover. Through the combination of these two modules, RiWFPN can obtain a “cleaner” feature map with better structure under adverse weather conditions. Next, we propose a moving object instance segmentation network with our proposed RiWFPN by extending SOLOv2~\cite{SOLOv2_2020}, called RiWNet, whose inputs are the pair of RGB frames. We design the ConvLSTM~\cite{ConvLSTM2015} based structure to introduce the temporal information of the next frame into current frame feature map, and guide the network to learn the motion information in the pair of frames. In the word, RiWNet is a novel moving object instance segmentation network being capable of obtain reliable and robust results in harsh weather disturbances. For training and evaluating the effectiveness of our method, we reorganize the VKITTI (Virtual KITTI)~\cite{VKITTI2020} dataset and change the original instance segmentation labels of all objects into instance segmentation labels of moving objects. In general, we propose VKITTI-moving dataset which is a moving instance segmentation dataset considering different weather conditions including rain, fog, sunset, morning as well as overcast, and is also divided into training and testing set manually.

%Next, we propose a moving object instance segmentation network with SOLOv2~\cite{SOLOv2_2020}, ConvLSTM~\cite{ConvLSTM2015} and our proposed RiWFPN, called RiWNet, whose inputs are the pair of RGB frames. 

In summary, the main contributions of this work are as follows: 
\begin{enumerate}
	\renewcommand{\labelenumi}{(\theenumi)}
	\item We propose RiWFPN including a progressive top-down interaction and attention refinement module to enhance the low-frequency structure information of the feature map. RiWFPN can improve the robustness and reliability of the network in varying adverse weather conditions after being directly inserted into the network as its neck structure instead of other FPN methods.
	\item We propose the RiWNet based on RiWFPN, which is a novel end-to-end moving object instance segmentation network being able to perform well in multiple severe weather conditions. RiWNet extends SOLOV2 with designed ConvLSTM-based structure to introduce temporal information to current feature map, and guide the network to learn motion information of the object in the pair of frames.
	\item To verify the effectiveness of our method, we propose a publicly available benchmark for moving instance segmentation, called VKITTI-moving dataset that takes into account weather conditions such as rain, fog, sunset, morning and overcast
	\item In the task of moving instance segmentation, the results have proved the ability and effect of RiWFPN to improve the robustness of the network in weather disturbance. The experimental results also show that the proposed RiWNet achieves state-of-the-art performance in some challenging datasets, especially under adverse weather scenarios.
\end{enumerate}

\begin{figure*}
	%	\vspace{-1.2cm}
	\begin{center}
		\includegraphics[width=1\linewidth]{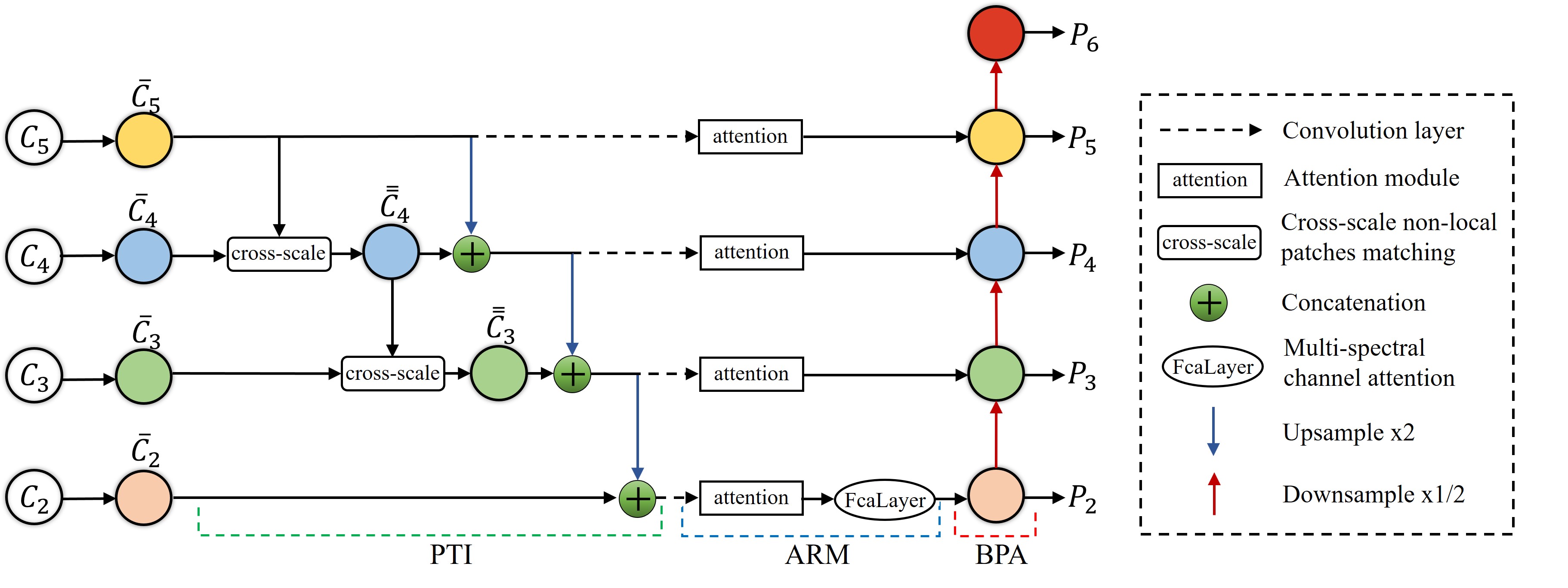}
	\end{center}
	\caption{Illustration of RiWFPN. RiWFPN includes progressive top-down interaction module (PTI), attention refinement module (ARM) and bottom-up path augmentation (BPA) three components.}
	\label{fig:fig1}
\end{figure*}

\section{Related work}

\subsection{Moving Object Segmentation}
The traditional multi-motion segmentation method~\cite{zhang2017permutation, xu20193d,zhao2019motion,ZhaoZQL20} using powerful geometric constraints cluster points of the scene into objects with different motion models. This type of method is at the feature point level instead of instance-level, limited by the model number of the segmentation and has a high computational burden. Some deep-learning-based~\cite{siam2018modnet,rashed2019fusemodnet,lu2019see,Automatic2020,Adversarial2020} methods can segment foreground moving objects from the dynamic scene without distinguishing each object instance. MODNet~\cite{siam2018modnet} proposes a novel two-stream architecture combining appearance and motion cues, and FuseMODNet proposes a real-time architecture fusing motion information
from both camera and LiDAR. More recent approaches~\cite{shen2018submodular,bideau2018best,xie2019object,dave2019towards,muthu2020motion,U2ONet2021} have used motion information from optical flow for the instance-level moving object segmentation. The method proposed in~\cite{xie2019object} discovers different moving objects based on their motion by foreground motion clustering. U$^2$-ONet~\cite{U2ONet2021} proposes a novel two-level nested U-structure to learn to segment moving objects and utilizes octave convolution (OctConv)~\cite{OctConv2019} to reduce computational burden. However, these methods are operated under in clear weather conditions, and they have problems inevitably under severe weather conditions. In contrast, RiWNet performs well even in severe weather environments.

\subsection{Robust Perception in Adverse Weather Conditions}
For robust perception in adverse weather environment, some common methods~\cite{VPGNet2017,PixelAccurate2019,RainRemoval2019,Semantic2019,CNNbased2020,ALittleFog2020WACV} try to obtain the data with various adverse weather effects, and then use these data to train the network to improve the influence of the network on real weather interference. The methods proposed in~\cite{Semantic2019} and~\cite{RainRemoval2019} propose a Foggy Cityscapes dataset with simulating fog and a RainCityscapes dataset with synthesizing rain streak respectively based on the generic Cityscapes~\cite{Cordts2016Cityscapes} dataset. This type of methods increases the amount of training data and greatly increases the training cost. Recently, some methods~\cite{NighttoDay2019,AutomaticAdaptation2019,CrossDomain2019,UnsupervisedMonocular2020,hnewa2021multiscale} regard each weather condition as a new domain, and improve the effect of the network in severe weather conditions based on domain adaptation methods. MS-DAYOLOs~\cite{hnewa2021multiscale} performs domain adaptation at multi-layer features from backbone network to generate domain invariant features for YOLOV4~\cite{YOLOv4}. Due to the addition of additional structures or features, these methods increase the difficulty of network training. Other common methods~\cite{RobustSemantic2019,RadarCamera2019,SeeingThrough2020} consider that different sensors perform differently in severe weather conditions and use multi-sensor fusion methods to improve the effect of perception. However, multi-sensor data is often difficult to obtain for cost or scenario constraints. Another possibility~\cite{pfeuffer2020robust} is to use the way of video processing and try to compensate the perception error of the current frame by using the image information of the sequence frame. The method proposed in~\cite{pfeuffer2020robust} modifies the recurrent units to ensure real-time performance and introduces a robust semantic segmentation using video-segmentation. In the proposed approach, we focus on a new possibility that is to make the network perform robustly in harsh weather conditions by improving its resilience to adverse weather effects through the network’s structural design.

\subsection{Architecture for Pyramidal Representations}
For deep learning-based perception tasks, features from different levels of the pyramid representation are often used. As a basic work, FPN (Feature Pyramid Network)~\cite{FPN2017} adopts a top-down pyramidal structure to represent multi-scale features. Taking FPN as a baseline, PANet~\cite{panet2018} creates a bottom-up path augmentation to much enhance FPN. Different from standard FPN, RFP (Recursive Feature Pyramid)~\cite{RFP2020} designs extra feedback connections into the bottom-up backbone layers. Feature Pyramid Grids (FPG)~\cite{FPG2020} represents the feature scale space as a regular grid that combines multi-directional horizontal connections and bottom-up parallel paths. NAS-FPN~\cite{NASFPN2019} combines neural architecture search to learn the optimal feature pyramid structure. HRFPN~\cite{sun2019deep,SunZJCXLMWLW19} concatenates the upsampled representations from all the backbone layers to augment the high-resolution features and uses average pooling to downsample the concatenated representation for constructing a multi-level representation. To deal with noisy images, OcSaFPN~\cite{OcSaFPN2020} improves the noise-resilient ability of the network itself by increasing the interaction between different frequency components and compressing the redundant information of low frequency components. However, these FPN methods do not consider to deal with the interference caused by bad weather conditions. By using RiWFPN as the neck structure instead of other FPN methods, the performance of our method in severe weather environments can be directly improved. 

\section{Method}
Firstly, robust RiWFPN is described, including how to improve the robustness of the network in the interference of weather conditions. Then, the overall structure of RiWNet is introduced, including its inputs and how to learn motion information. The proposed moving instance segmentation dataset considering various weather conditions is introduced at the end of this section. 

\subsection{RiWFPN}   
\label{sec:RiWFPN}
\subsubsection{overview}
As shown in Fig.~\ref{fig:fig1}, inputs of RiWFPN are the feature maps of four levels $\left\{C_2,C_3,C_4,C_5 \right\}$ generated by the backbone. It has been demonstrated in the literature that noise in features can be drastically reduced via re-scaling to coarser pyramid level and noisy patches as well as edge patches usually have their corresponding "clean" patches at coarser image scales of the same relative image coordinates~\cite{Separating2013}. Inspired by this conclusion, we first perform a progressive top-down interaction module to enhance the feature map of each scale. This module borrows low-scale “clean“ information through cross-scale non-local patches matching, and increases the spread of clean information while introducing more spatial and semantic information through the concatenation of adjacent-scale feature maps. And then we use an attention refinement module to refine the abundant information for highlighting significant features for specific scales, and to enhance low-frequency structural information. Finally, the bottom-up path augmentation is used to strengthen the propagation of high-scale refined feature maps and optimize the feature maps of the entire network. 

\subsubsection{progressive top-down interaction module}
Inspired by~\cite{Separating2013} and~\cite{mei2020pyramid}, cross-scale non-local patches matching of adjacent scales is used to introduce “clean” information on lower scales (especially better edge information) into higher scales. Cross-scale non-local patches matching is first performed on feature maps of adjacent scales $\left\{\overline{C}_4,\overline{C}_5\right\}$, bringing “clean” recurrence information on lower scales into current scales to obtain $\left\{\overline{\overline{C}}_4\right\}$ which is the "cleaner" version of $\left\{C_4\right\}$. Recursively, the feature map $\left\{\overline{\overline{C}}_4\right\}$ is then subjected to cross-scale non-local patches matching operation with $\left\{\overline{C}_3\right\}$. The feature map $\left\{\overline{C}_2\right\}$ with the highest scale is larger in size. In order to avoid adding too much computational overhead, no cross-scale matching operation is performed on $\left\{\overline{C}_2,\overline{\overline{C}}_3\right\}$. 

Formally, given two input feature maps $F$ and $G$ of adjacent scales (the scale of $F$ is greater than $G$), cross-scale non-local patches matching operation is defined as: 
\begin{equation}\label{eq1}
y^i = \frac{1}{\sigma(F,G)}\sum_{j}\phi(F^i_{\delta(r)},G^j_{\delta(r)})\theta(G^j),
\end{equation}
where $i$, $j$ are index on the input $F$ and input $G$ as well as output $y$. The function $\phi$ computes pair-wise affinity between two input features. $\theta$ is a feature transformation function that generates a new representation of $G^j$. The output response $y^i$ obtains information from all features through explicitly summing over all positions and is normalized by a scalar function $\sigma(F,G)$. The neighborhood is specified by $\delta(r)$. $r \times r$ patches are extracted from the feature map.
For $\phi$, we use embedded Gaussian\cite{liu2018non} as:
\begin{equation}\label{eq2}
\phi(F^i,G^j) = e^{f(F^i)^Tg(G^j)},
\end{equation}
The scalar function $\sigma(F,G)$ is set as:
\begin{equation}\label{eq3}
\sigma(F,G) = \sum_{j\in G}\phi(F^i,G^j)
\end{equation}
where, $f(F^i) = W_fF^i$ and $g(G^j) = W_gG^j$.
we use a simple linear embedding for the function $\theta$: $\theta = W_{\theta}G^j$. 

Then, the progressive top-down concatenation is performed. For each scale except the lowest scale of the pyramid, the current scale feature map and upsampled feature maps from its previous scale are concatenated. Through this progressive concatenation method, more low-level features are integrated and the fusion of features of different scales is promoted, to obtain more spatial and semantic information. At the same time, this concatenation interaction between different scales improves the propagation of “cleaner” feature maps obtained by cross-scale patch matching, and optimizes feature maps at various scales. Inspired by the conclusion proved by~\cite{OcSaFPN2020} that the transmission of the information at different frequency components can enhance noise-resilient performance of the network, this top-down progressive concatenation is also used to increase the interaction between different scales. In this way, the network's resilience to noise such as rain and fog is improved.

\subsubsection{attention refinement module}
\begin{figure}[t]
	\begin{center}
		\includegraphics[width=1.0\linewidth]{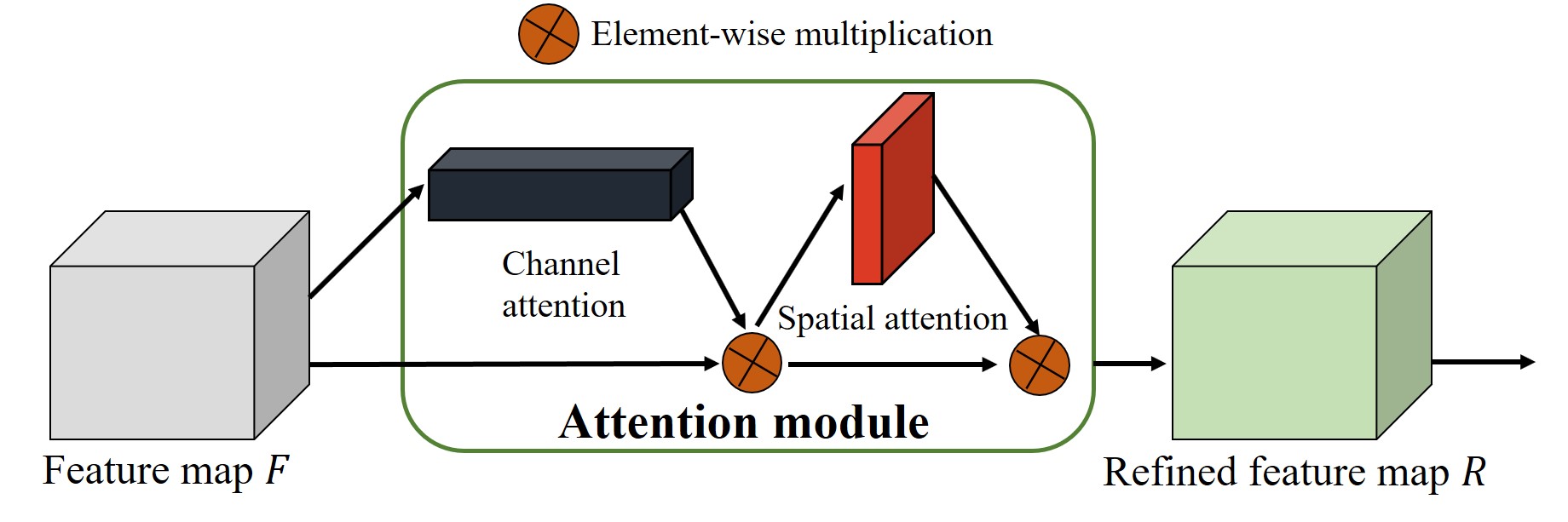}
	\end{center}
	\caption{Overview of convolutional block attention module (CBAM)~\cite{woo2018cbam}. CBAM introduces channel and spatial attention mechanisms.}
	\label{fig:fig2}
\end{figure}

The abundant feature maps after concatenation have vast spatial and channel aggregation information of multi-scale concatenated feature maps, but making scale features of targets not significant enough. For feature map $\left\{\overline{C}_5\right\}$ and the three feature maps after concatenation, convolutional block attention module (CBAM)~\cite{woo2018cbam} is used to fuse multi-scale feature maps for highlighting significant features of specific scales. CBAM (as shown in Fig.~\ref{fig:fig2}) weights in dimensions of scale and spatial respectively by introducing channel and spatial attention mechanisms. Channel attention attempts to selecting feature maps of suitable scales and spatial attention concentrates on finding salient portions in a feature map. Through this operation, multiscale feature maps are refined adaptively by CBAM, to emphasize the prominent features of specific scales and pay more attention to specific scales for multi-scale segmentation.

\begin{figure*}
	\begin{center}
		\includegraphics[width=1\linewidth]{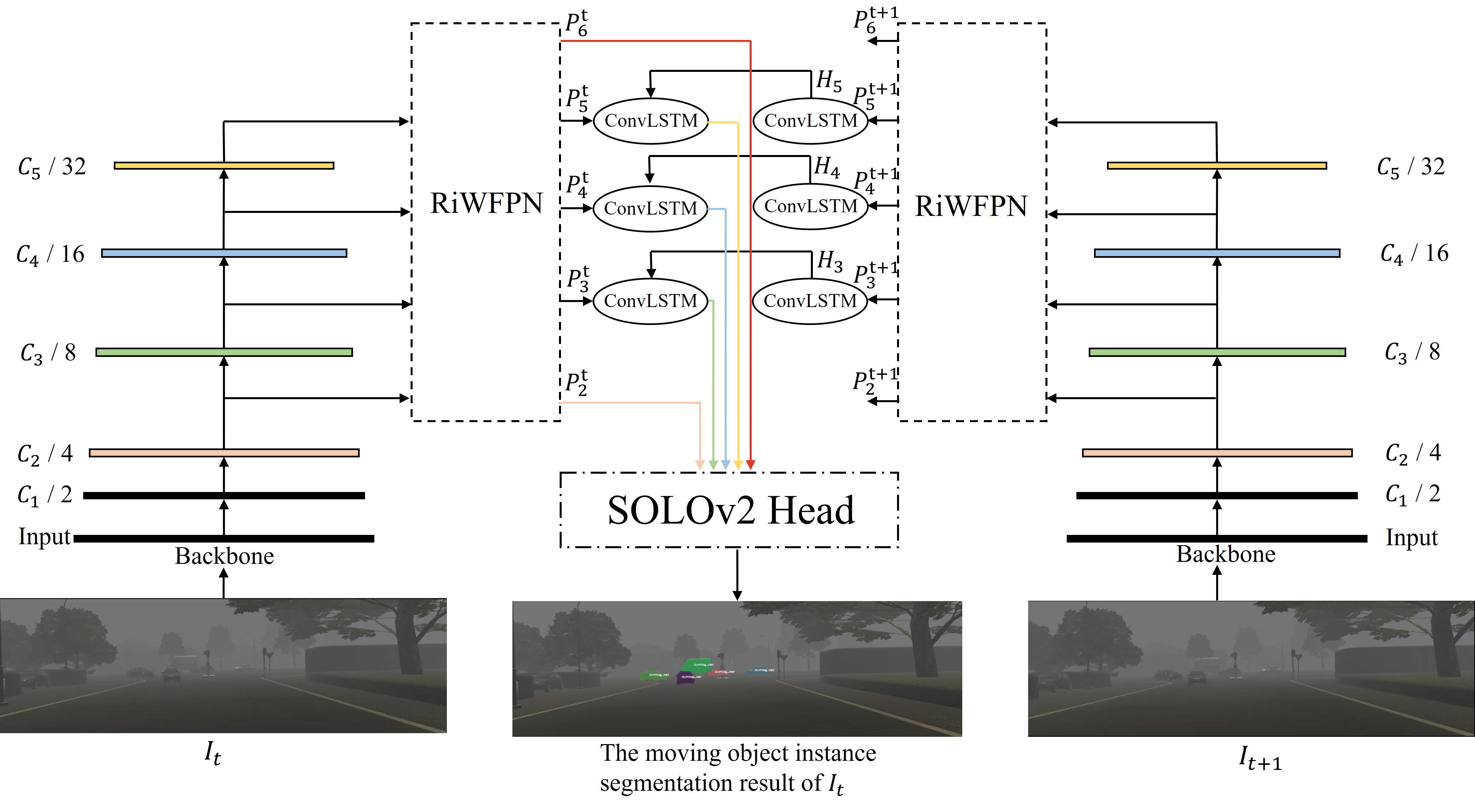}
	\end{center}
	\caption{Overview of RiWNet. The input of RiWNet is two adjacent frames of RGB images $I_t,I_{t+1}$, and the output is the segmentation result of the moving object instance in the current frame $I_t$. The RiWFPN used is introduced in Section~\ref{sec:RiWFPN}.}
	
	\label{fig:fig3}
\end{figure*}

The existing attention model using GAP as pre-processing method only uses the information equivalent to the lowest frequency of the DCT (Discrete Cosine Transform), while discarding much useful information equivalent to other frequency of the DCT, as mentioned in~\cite{fcanet2020}. Therefore, the multi-spectral attention module Fcalayer~\cite{fcanet2020} is proposed to exploit the information from different frequency components of the DCT, to making full use of the information in the attention mechanism more efficiently. Based on the conclusion of~\cite{Separating2013} that noise levels drop dramatically at coarser image scales, we use Fcalayer to optimize the highest-scale feature map $\left\{C_2\right\}$ that we think is the most noise levels. Fcalayer embeds different frequency information, and selects Top-k highest performance frequency components in each frequency. In addition, the frequency components selected by the Fcalayer selection mechanism is usually more biased towards selecting low-frequency, as proved in~\cite{fcanet2020}. Therefore, the highest-scale feature map optimized by Fcalayer can have rich low-frequency information to improve the structural information of the network. Better structural information helps improve the network's resilience to the weather interference such as rain and fog. 

\subsection{RiWNet} 
\label{sec:RiWNet}
As illustrated in Fig.~\ref{fig:fig3}, the overall structure of RiWNet is introduced. First, given a pair of RGB frames at adjacent time $I_t,I_{t+1} \in \mathbb{R}^{h\times w\times 3}$, we take ResNet101~\cite{resnet2016} as the backbone. $\left\{P_2^t,P_3^t,P_4^t,P_5^t,P_6^t\right\}$ and $\left\{P_2^{t+1},P_3^{t+1},P_4^{t+1},P_5^{t+1},P_6^{t+1}\right\}$ respectively represent feature levels generated by the proposed RiWFPN (in Section~\ref{sec:RiWFPN}) at time $t$ and $t+1$. Then the three feature maps $\left\{P_3^{t+1},P_4^{t+1},P_5^{t+1}\right\}$ at time $t+1$ are respectively input into a ConvLSTM structure~\cite{ConvLSTM2015}, and three feature maps $\left\{H_3,H_4,H_5\right\}$ are output. Considering that the large size of $\left\{P_2^{t+1}\right\}$ bring a lot of computational burden, and that $\left\{P_6^{t+1}\right\}$ contains less information, the feature maps of three levels $\left\{P_3^{t+1},P_4^{t+1},P_5^{t+1}\right\}$ are selected. At the same time, we also experimentally proved (in Section~\ref{subsubsec:RiWNet}) that using the feature maps of three levels $\left\{P_3^{t+1},P_4^{t+1},P_5^{t+1}\right\}$ is better than using only the feature map $\left\{P_2^{t+1}\right\}$. Three feature maps $\left\{P_3^t,P_4^t,P_5^t\right\}$ and three feature maps $\left\{H_3^t,H_4^t,H_5^t\right\}$ at time $t+1$ are respectively used as the input feature map and hidden layer feature map of the ConvLSTM structure for processing. As mentioned in ~\cite{pfeuffer2020robust}, ConvLSTMs capture the temporal image information well. In this way, the ConvLSTM is used to introduce the information at time $t+1$ into the feature map at time $t$ to realize the processing of temporal information and guide the network to learn the motion information in adjacent temporal images in addition to appearance information. Finally, the three feature maps after processed by the ConvLSTM structure and the two feature maps $\left\{P_2^t,P_6^t\right\}$ at time $t$ are used as the input of SOLOv2 Head~\cite{SOLOv2_2020} to obtain the result of moving instance segmentation. 

\subsection{VKITTI-moving} 
\label{sec:VKITTI-moving}
\begin{figure*}
	\subfigure[] {
		\label{fig:fig4-a}     
		\includegraphics[width=0.33\linewidth]{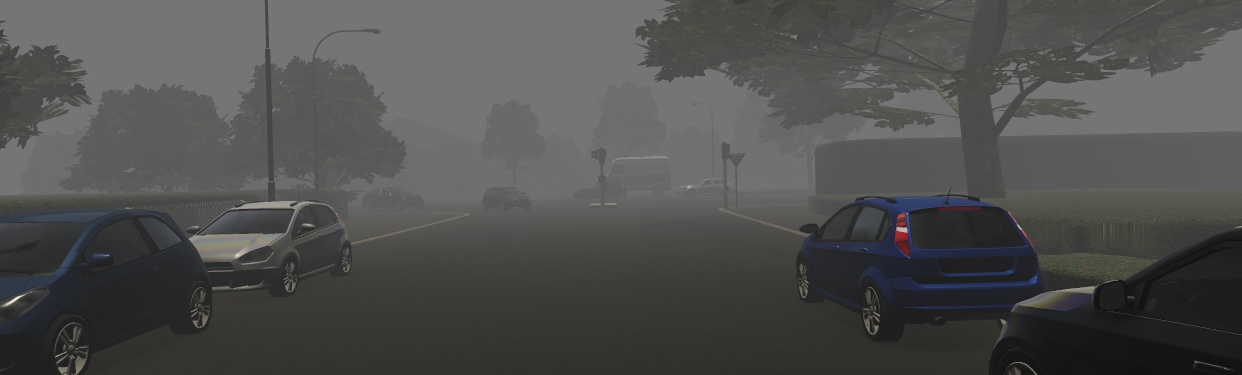}  
	} 
	\subfigure[] {
		\label{fig:fig4-b}     
		\includegraphics[width=0.33\linewidth]{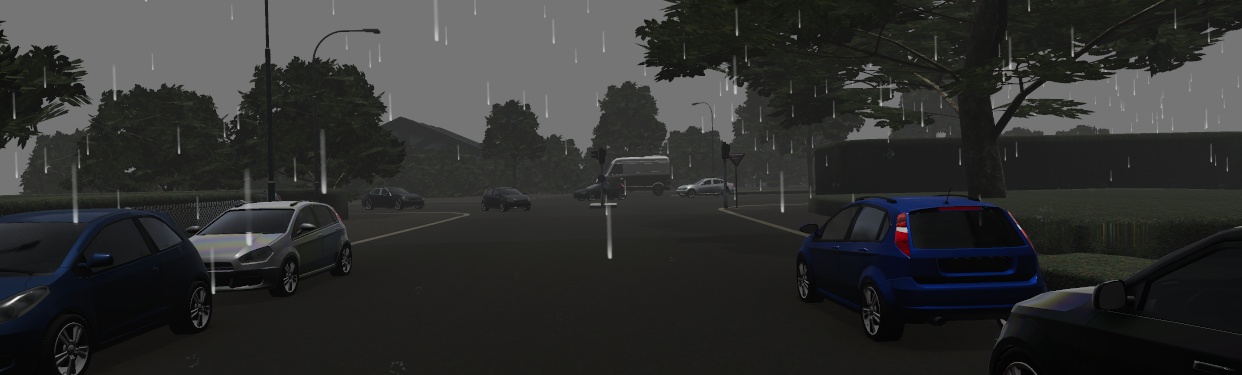}  
	} 
	\subfigure[] {
		\label{fig:fig4-c}     
		\includegraphics[width=0.33\linewidth]{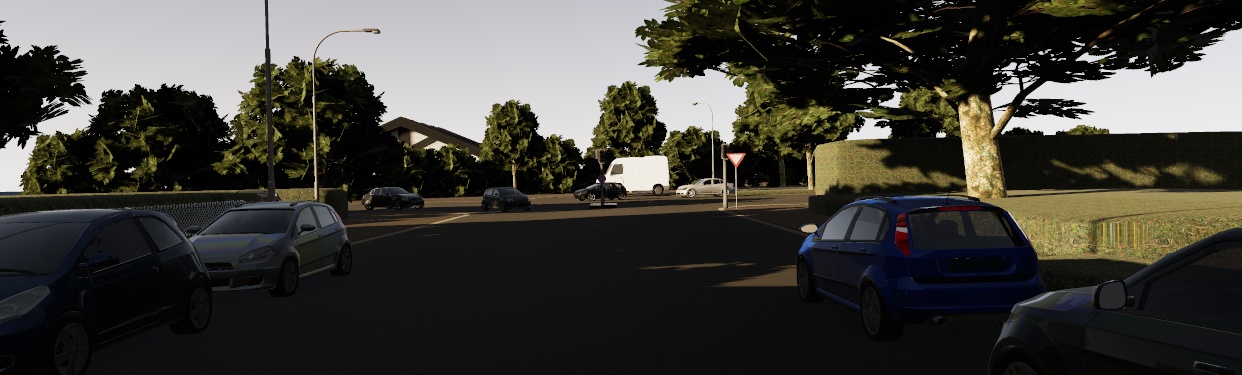}  
	} 
	\subfigure[] {
		\label{fig:fig4-d}     
		\includegraphics[width=0.33\linewidth]{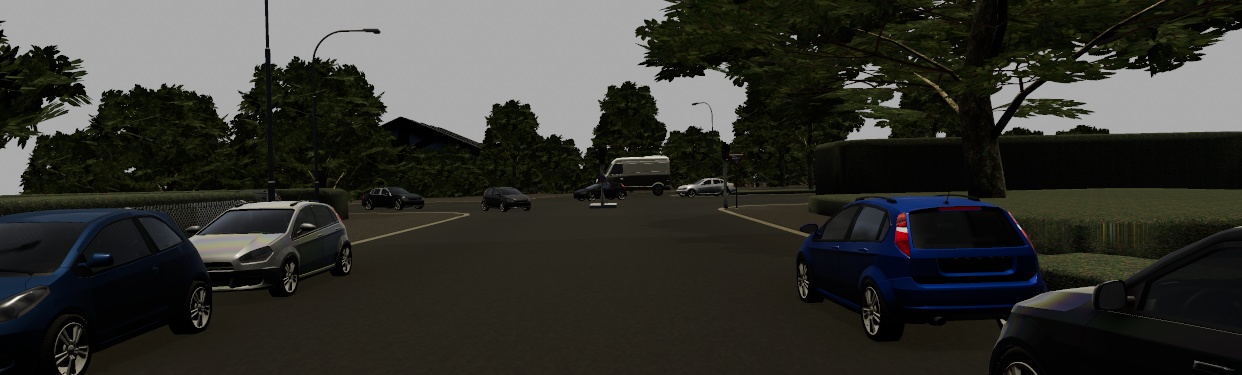} 
	} 
	\subfigure[] {
		\label{fig:fig4-e}     
		\includegraphics[width=0.33\linewidth]{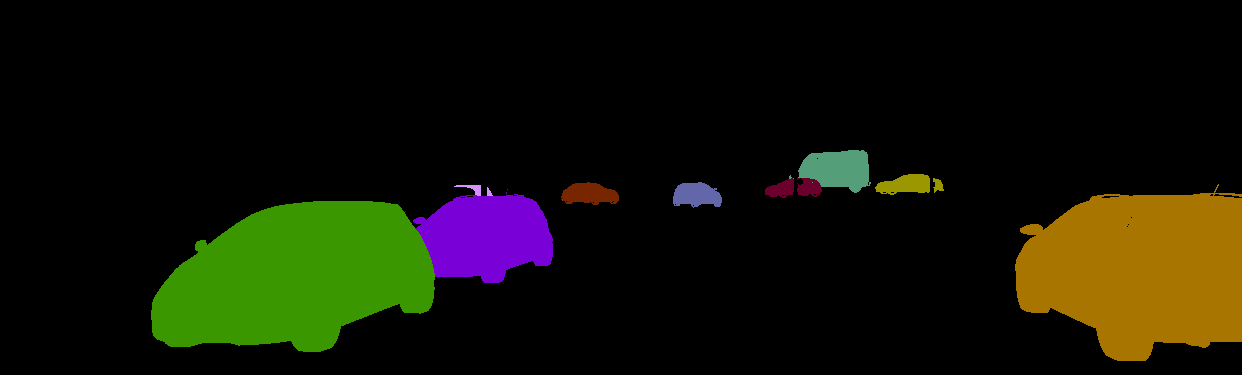}  
	} 
	\subfigure[] {
		\label{fig:fig4-f}     
		\includegraphics[width=0.33\linewidth]{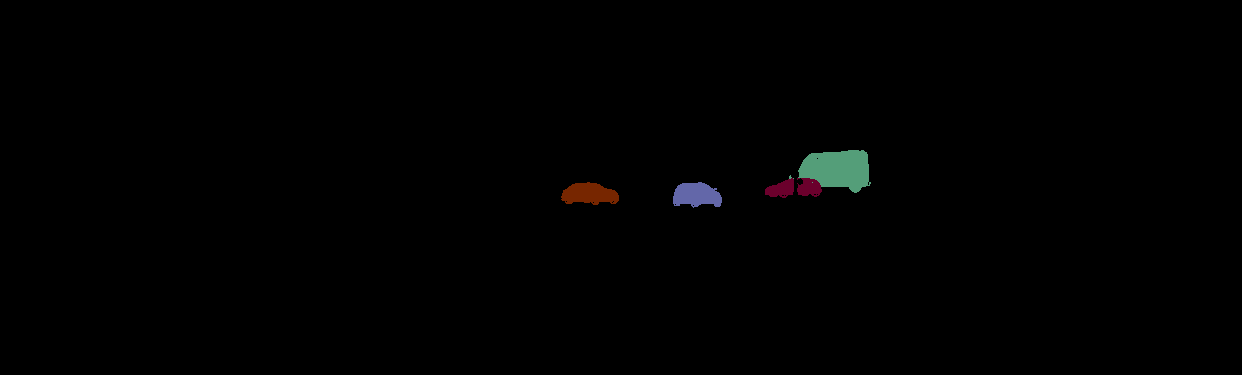}  
	} 
	\caption{The example image of VKITTI-moving. (a) Fog image. (a) Rain image. (c) Sunset image. (d) Overcast image. (e) Instance segmentation labels of VKITTI. (f) Moving instance segmentation labels of our VKITTI-moving.}
	\label{fig:fig4}
\end{figure*}

To train and verify our deep model for robust performance in weather disturbances, we propose a publicly available moving instance segmentation dataset, called VKITTI-moving by generating motion masks annotations from a large driving dataset-VKITTI (Virtual KITTI)~\cite{VKITTI2020} (as shown in Fig.~\ref{fig:fig4}). VKITTI dataset consists of 5 tracking sequence and provides these sequences after modified weather conditions (e.g. fog, rain, sunset, morning and overcast). Although VKITTI contains left and right camera images under different camera configurations (e.g. rotated by $30\degree$), they are all images from different perspectives collected in the same scene, we only take "15-deg-left-Camera\_0" collected by the left camera rotated 15 degrees. To generate motion masks from VKITTI, we take the instance segmentation groundtruth mask, manually select and retain masks belonging to moving objects. In addition to the background, VKITTI-moving has only one category: moving\_car, the category of all labeling instance masks is moving\_car. VKITTI-moving considers different weather conditions including rain, fog, sunset, morning and overcast in the same scene. It contains 4650 images (shown in TABLE~\ref{tab:dataset}), and is divided to three subsets: Fog, Rain and Illumination (including sunset, morning and overcast). The size of the image is (1242, 375). For quantifying the performance, we use the precision (P), recall (R), and F-measure(F), as defined in~\cite{SegmentationAnalysis2014}, as well as the mean intersection over union (IoU) for the evaluation metrics. 

\begin{table}[htbp]
  \scriptsize	
  \centering
  \caption{Introduction of the VKITTI-moving dataset}
    \begin{tabular}{ccccc}
    \toprule
    Subset &       & \multicolumn{1}{p{7.065em}}{Image quantity in training set} & \multicolumn{1}{p{7.065em}}{Image quantity in testing set} & \multicolumn{1}{p{4.825em}}{Instances} \\
    \midrule
    Fog   &       & 660   & 270   & 2550 \\
    \midrule
    Rain  &       & 660   & 270   & 2550 \\
    \midrule
    \multirow{3}[0]{*}{Illumination} & sunset & 660   & 270   & 2550 \\
          \cmidrule(r){2-5}
          & morning & 660   & 270   & 2550 \\
          \cmidrule(r){2-5}
          & overcast & 660   & 270   & 2550 \\
    \bottomrule
    \end{tabular}%
  \label{tab:dataset}%
\end{table}%

\section{Experiments}
\subsection{Datasets}
We evaluated the proposed method on several benchmark datasets: our VKITTI-moving (in Section~\ref{sec:VKITTI-moving}), FBMS (Freiburg–Berkeley Motion Segmentation) dataset~\cite{SegmentationAnalysis2014}, YTVOS (YouTube Video Object Segmentation) dataset~\cite{ytvos2018}. FBMS is a widely used moving object segmentation dataset, and many methods are tested on this dataset. We used a corrected version~\cite{DetailedRubric2016} linked from FBMS’s website because the original FBMS has a lot of annotation errors. The YTVOS dataset is a challenging video object segmentation dataset containing many objects that are difficult to segment, such as tiny objects, and camouflaged objects. For testing moving object segmentation, we used the moving object version of YTVOS, called YTVOS-moving proposed in ~\cite{dave2019towards}. 

\subsection{Implementation Details}
For the experiments, we take ResNet101 pretrained on ImageNet~\cite{ImageNet2009} as the backbone. As for VKITTI-moving dataset, the longer image side is 1242. We use scale jitter for the shorter image side, and it is randomly sampled from 800 to 640 pixels. For FBMS and YTVOS, the longer image side is set to 1242, and the shorter image side is randomly sampled from 512 to 352 pixels. RiWNet is trained with stochastic gradient descent (SGD). Its initial learning rate is 0.001 and its hyperparameters are set as follows: momentum=0.9, weight\_decay=0.00001. We train for 40 epochs using a batch size of 2. The experiments are all conducted on a single NVIDIA Tesla V100 GPU with 16GB memory, along with the PyTorch 1.4.0 and Python 3.7. In the VKITTI-moving dataset, RiWNet can run end-to-end at a speed of about 5HZ in this configuration. The results are evaluated using the mean intersection over union (IoU), precision (P), recall (R), and F-measure (F). Source code and the models as well as VKITTI-moving dataset will be made public in \emph{\textcolor[rgb]{1,0,0}{\url{https://github.com/ChenjieWang/RiWNet}}} upon the publishment of the paper.

\subsection{Visualization of Feature Maps}
\begin{figure*}
	\subfigure[] {
		\label{fig:fig5-a}     
		\includegraphics[width=0.5\linewidth]{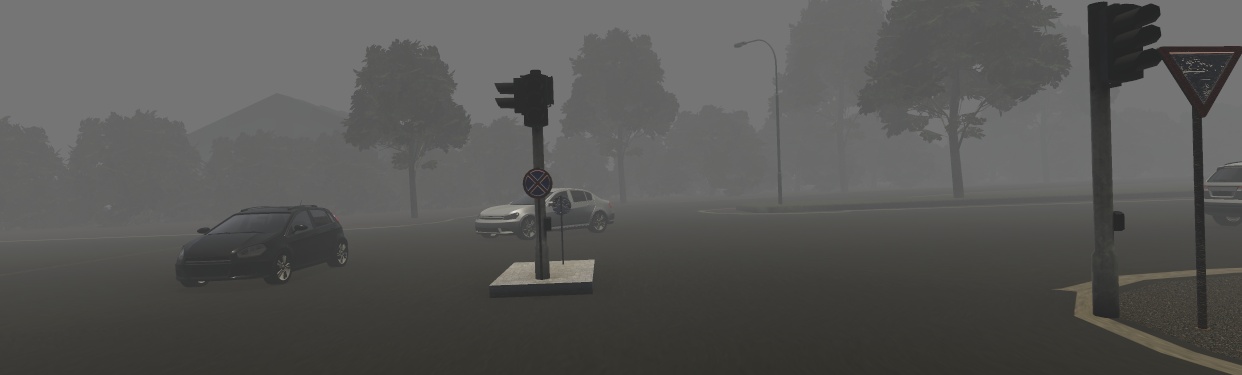}  
	} 
	\subfigure[] {
		\label{fig:fig5-b}     
		\includegraphics[width=0.5\linewidth]{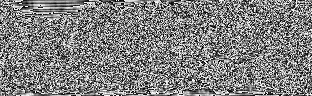}
	} 
	\subfigure[] {
		\label{fig:fig5-c}     
		\includegraphics[width=0.5\linewidth]{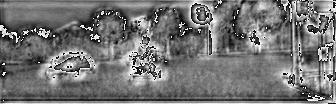} 
	} 
	\subfigure[] {
		\label{fig:fig5-d}     
		\includegraphics[width=0.5\linewidth]{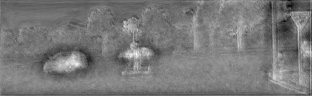}
	} 
	\caption{Comparison of visualization of the mean feature map obtained using different FPN structures. (a) Original image with fog. (a) The mean feature map obtained with NASFPN. (c) The mean feature map obtained with HRFPN. (d) The mean feature map obtained with RiWFPN.}
	\label{fig:fig5}
\end{figure*}

To fully demonstrate how RiWNet enhance the feature map, we visualize related processing results of the feature map $\left\{P_2^t\right\}$ in our method. First, we average the values of all channels on each pixel coordinate (i, j) to obtain the mean feature map. For comparison, we also calculated the mean feature map of RiWNet using other FPN structures (including HRFPN and NASFPN) instead of RiWFPN. The quantitative results (in Section~\ref{subsubsec:ComparisonFPN}) prove that these two FPN structures are the best two structures in weather conditions other than RiWFPN, so we compare our RiWFPN with these two structures. We visualize these mean feature maps in Fig.~\ref{fig:fig5}. In comparison, the feature maps of all channels of our RiWFPN are more consistent, the mean feature map has better low-frequency structure information, and the target object is more significant.

\begin{figure}[t]
	\begin{center}
		\includegraphics[width=1.0\linewidth]{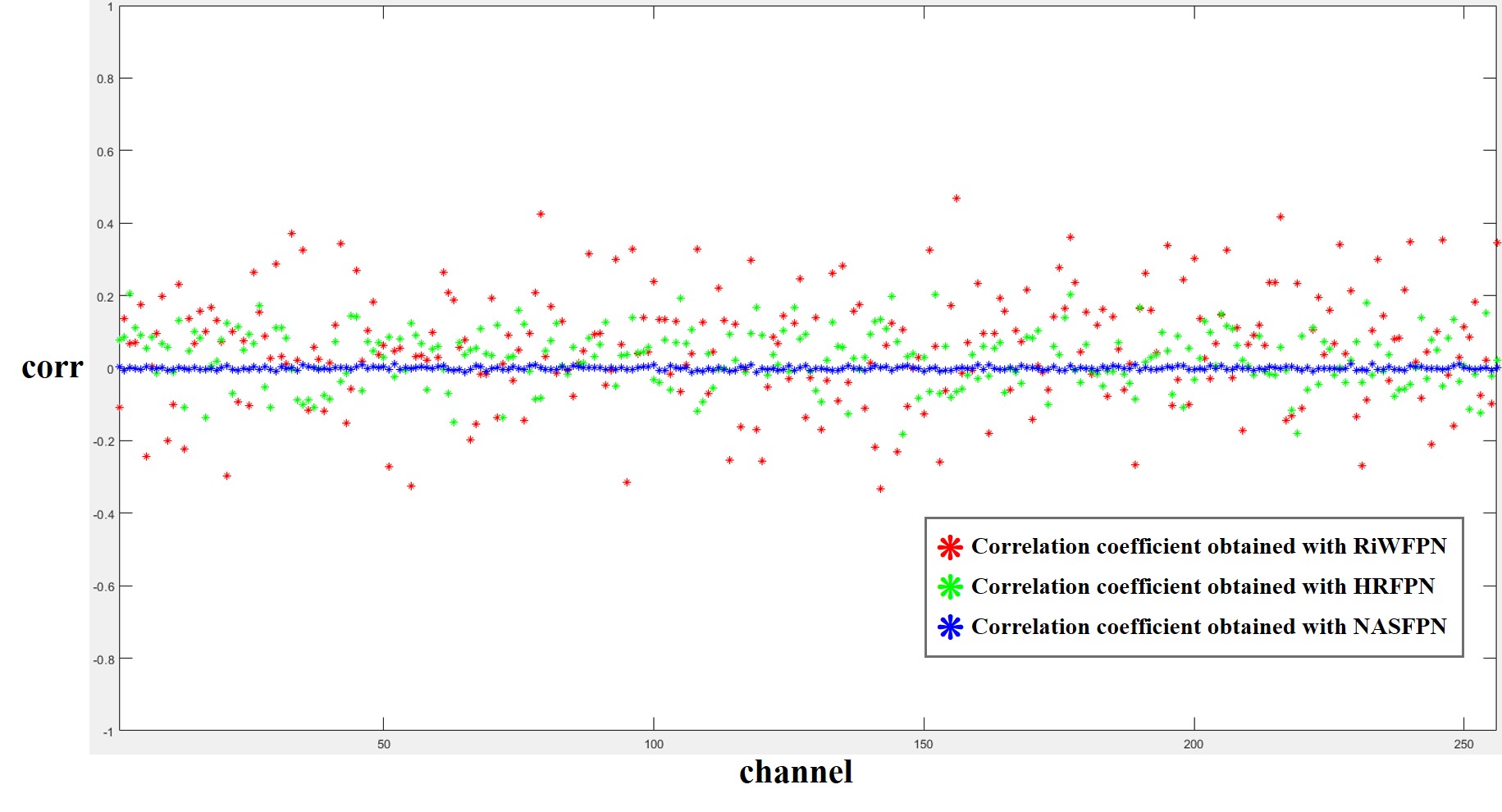}
	\end{center}
	\caption{Comparison of correlation coefficients obtained using different FPN structures. The correlation coefficient distribution of RiWFPN is more discrete.}
	\label{fig:fig6}
\end{figure}

\begin{figure*}
	\subfigure[] {
		\label{fig:fig7-a}     
		\includegraphics[width=0.30\linewidth]{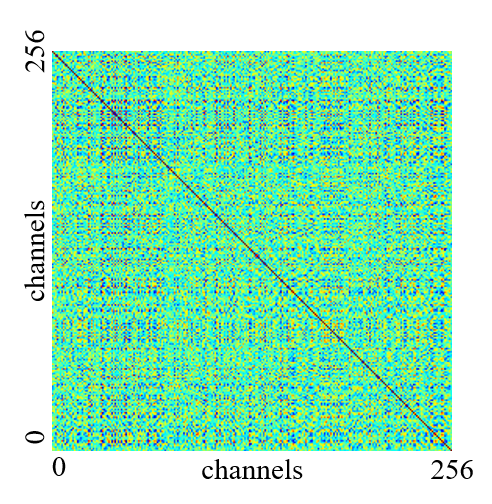}  
	} 
	\hspace{-0.2cm}
	\subfigure[] {
		\label{fig:fig7-b}     
		\includegraphics[width=0.30\linewidth]{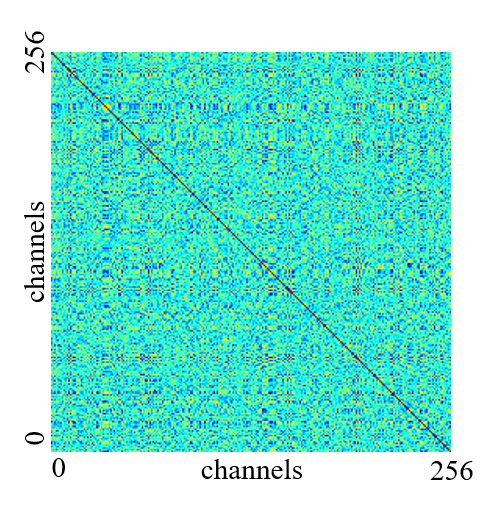}  
	} 
	\hspace{-0.2cm}
	\subfigure[] {
		\label{fig:fig7-c}     
		\includegraphics[width=0.30\linewidth]{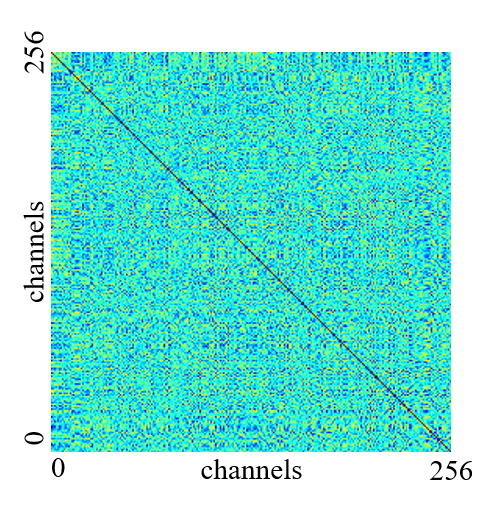}  
	} 
	\hspace{-0.2cm}
	\subfigure[] {
		\label{fig:fig7-d}     
		\includegraphics[width=0.05\linewidth]{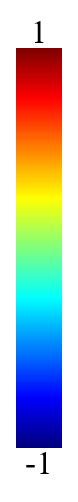}  
	} 
	\caption{(a) Visualized image of Correlation coefficient of NASFPN. (b) Visualized image of Correlation coefficient of HRFPN. (c) Visualized image of Correlation coefficient of RiWFPN. (d) Colormap indicating that the correlation coefficient is from -1 to 1.}
	\label{fig:fig7}
\end{figure*}

At the same time, in order to prove that the consistency of our method is not due to the decrease in the amount of information in each channel and the increase in the duplication of information in each channel. We calculate the correlation coefficient between the feature map of each channel and the mean feature map respectively. Meanwhile, we also calculate the correlation coefficient between every two channels in all channels. The correlation coefficient $corr$ is defined as:
\begin{equation}\label{eq4}
corr = \frac{\sum_{i}\sum_{j}(F^n_{ij}-\overline{F^n})(mF_{ij}-\overline{mF})}
{\sqrt{(\sum_{i}\sum_{j}(F^n_{ij}-\overline{F^n})^2)(\sum_{i}\sum_{j}(mF_{ij}-\overline{mF})^2)}}.
\end{equation}
The comparison of correlation coefficients are shown in Fig.~\ref{fig:fig6} and Fig.~\ref{fig:fig7}. It is interesting to point out, the correlation coefficient between each channel and the average channel or between every two channels of RiWFPN is generally lower. It indicates that the use of RiWFPN introduces more information. Combined with the results in Fig.~\ref{fig:fig5}, RiWFPN has introduced more information while enhancing the consistency of information in different channels. In general, RiWFPN enhances the low-frequency structure information in the network, which is more conducive to maintaining robustness in adverse weather interference. 
%This is being said

\begin{figure*}
	\subfigure[] {
		\label{fig:fig8-a}     
		\includegraphics[width=0.245\linewidth]{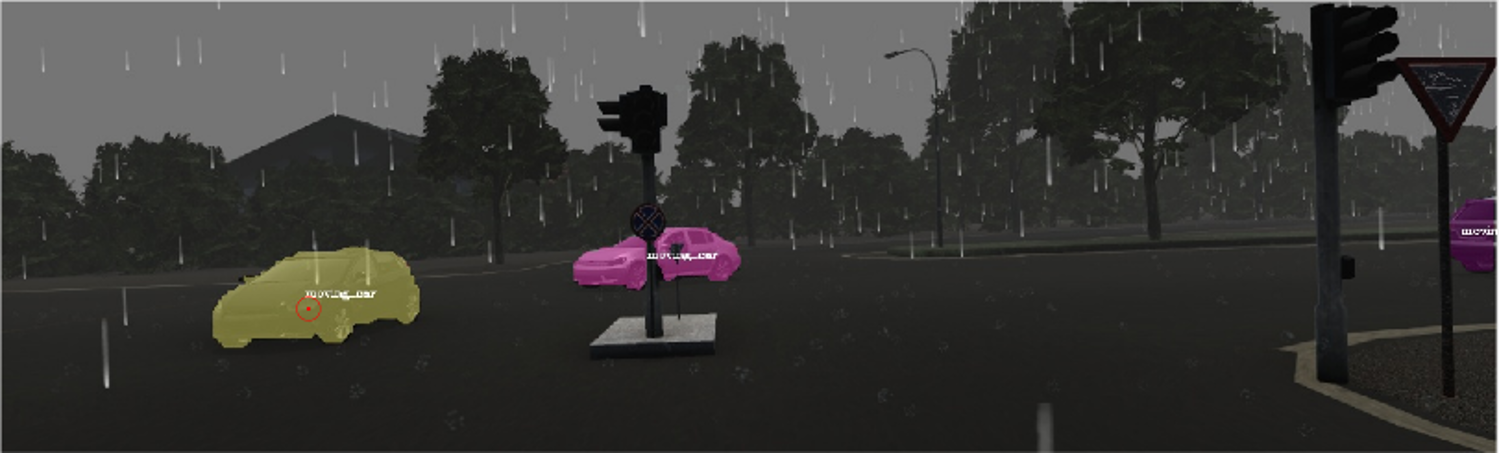}  
	} 
	\hspace{-1.0cm}
	\subfigure[] {
		\label{fig:fig8-b}     
		\includegraphics[width=0.245\linewidth]{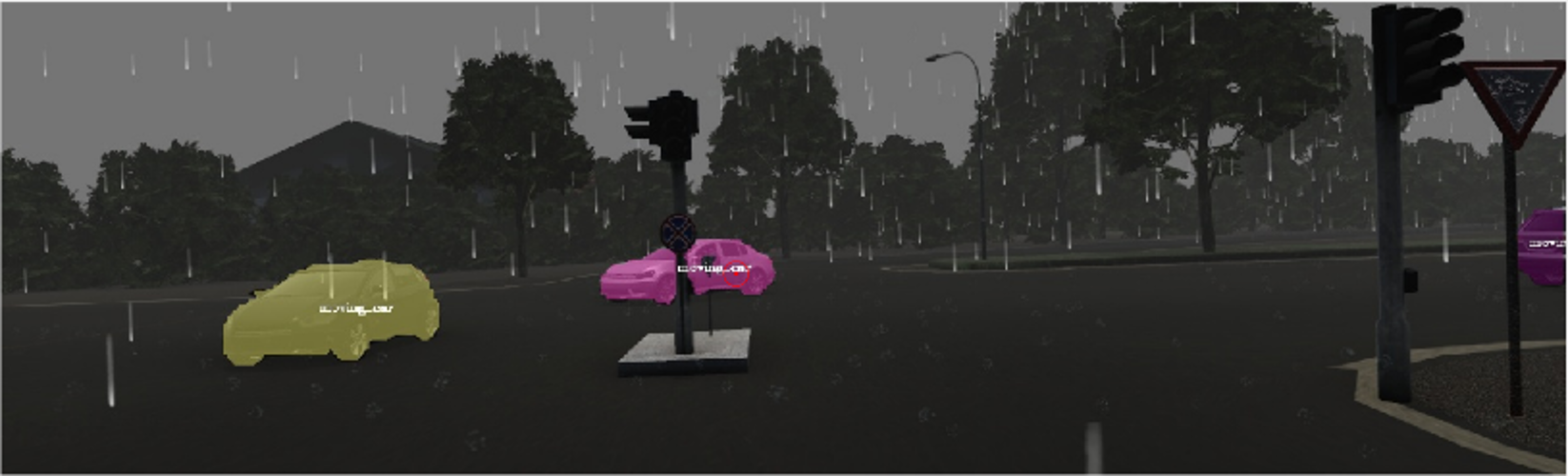}  
	} 
	\hspace{-1.0cm}
	\subfigure[] {
		\label{fig:fig8-c}     
		\includegraphics[width=0.245\linewidth]{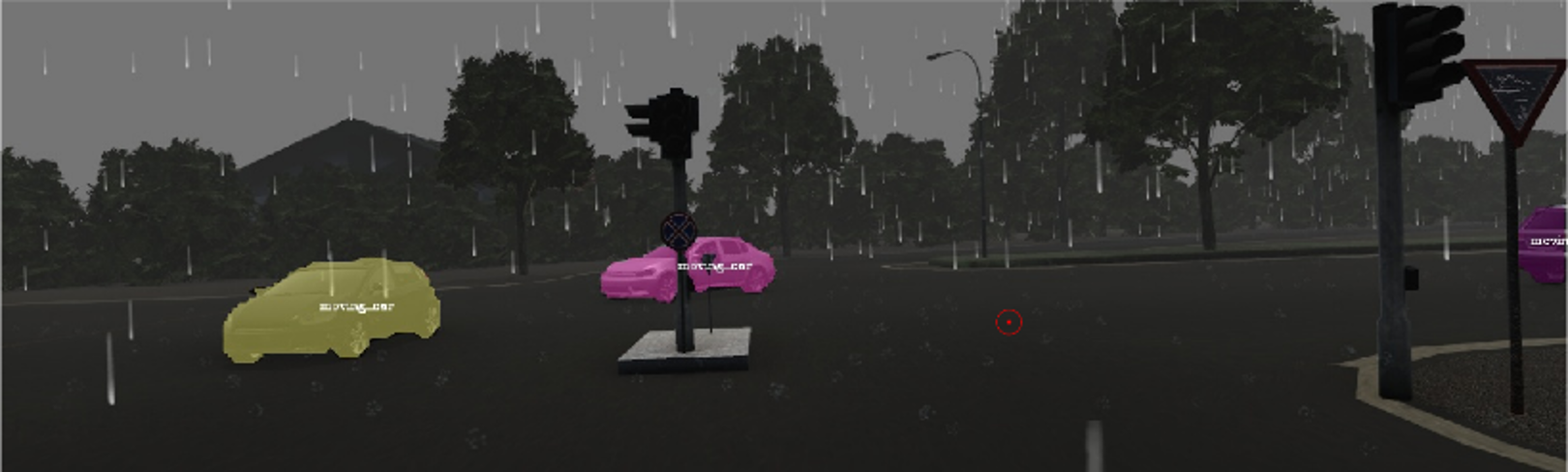}  
	} 
	\hspace{-1.0cm}
	\subfigure[] {
		\label{fig:fig8-d}     
		\includegraphics[width=0.245\linewidth]{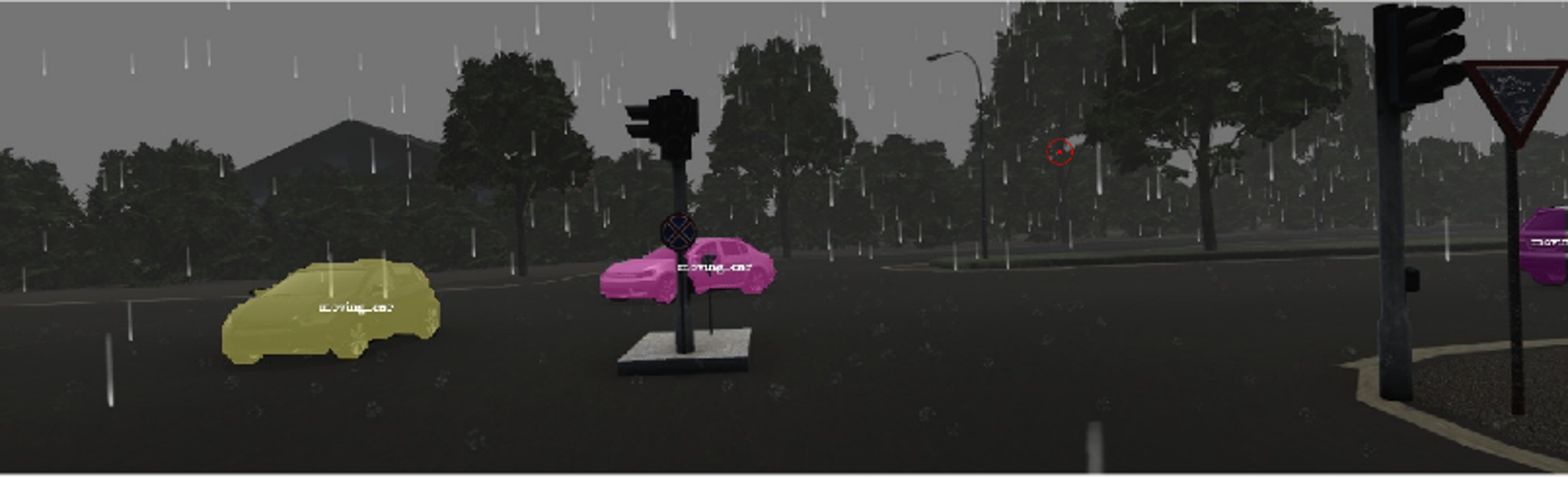} 
	} 
	\hspace{-1.0cm}
	\subfigure[] {
		\label{fig:fig8-e}     
		\includegraphics[width=0.245\linewidth]{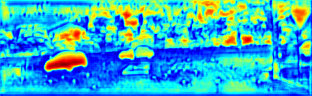}  
	} 
	\hspace{-1.0cm}
	\subfigure[] {
		\label{fig:fig8-f}     
		\includegraphics[width=0.245\linewidth]{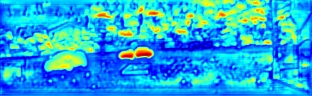}  
	} 
	\hspace{-1.0cm}
	\subfigure[] {
		\label{fig:fig8-g}     
		\includegraphics[width=0.245\linewidth]{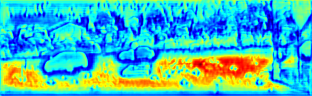}  
	} 
	\hspace{-1.0cm}
	\subfigure[] {
		\label{fig:fig8-h}     
		\includegraphics[width=0.245\linewidth]{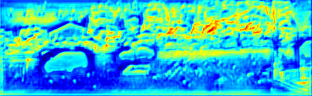} 
	}
	\hspace{-1.0cm}
	\subfigure[] {
		\label{fig:fig8-i}     
		\includegraphics[width=0.245\linewidth]{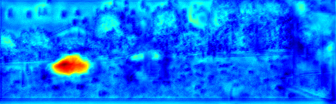}  
	} 
	\hspace{-1.0cm}
	\subfigure[] {
		\label{fig:fig8-j}     
		\includegraphics[width=0.245\linewidth]{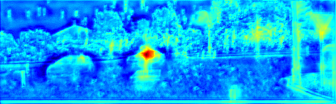}  
	} 
	\hspace{-1.0cm}
	\subfigure[] {
		\label{fig:fig8-k}     
		\includegraphics[width=0.245\linewidth]{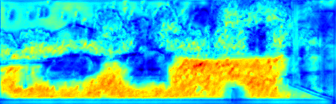}  
	} 
	\hspace{-1.0cm}
	\subfigure[] {
		\label{fig:fig8-l}     
		\includegraphics[width=0.245\linewidth]{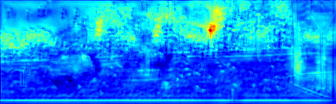} 
	} 
	\hspace{-1.0cm}
	\subfigure[] {
		\label{fig:fig8-m}     
		\includegraphics[width=0.245\linewidth]{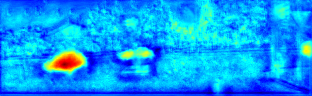}  
	} 
	\hspace{-0.4cm}
	\subfigure[] {
		\label{fig:fig8-n}     
		\includegraphics[width=0.245\linewidth]{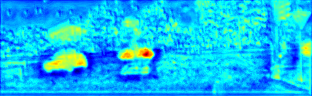}  
	} 
	\hspace{-0.4cm}
	\subfigure[] {
		\label{fig:fig8-o}     
		\includegraphics[width=0.245\linewidth]{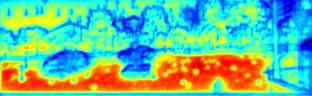}  
	} 
	\hspace{-0.4cm}
	\subfigure[] {
		\label{fig:fig8-p}     
		\includegraphics[width=0.245\linewidth]{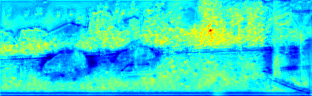} 
	} 
%	\subfigure[] {
%		\label{fig:fig4-q}     
%		\includegraphics[width=1\linewidth]{./figures/colorscale_jet.jpg}  
%	} 
	\caption{The correlation coefficient between the feature vector of each pixel coordinate and the reference pixel coordinate feature vector. The reference pixel position is the center of the red circle in the figure. The reference pixel coordinates in (a) and (b) are located on the target moving object. The reference pixel coordinates in (c) and (d) are located on the background. (a-d) Original image with rain. (e-h) Visualized image of Correlation coefficient of NASFPN. It can be seen that a large number of backgrounds are highly correlated with target objects. NASFPN has good intra-class correlation, but poor inter-class differences. (i-l) Visualized image of Correlation coefficient of HRFPN. It can be seen that the correlation within the target object class is not high. HRFPN has good inter-class differences, but poor intra-class correlation. (m-p) Visualized image of Correlation coefficient of RiWFPN. RiWFPN has good intra-class correlation and inter-class differences.}
	\label{fig:fig8}
\end{figure*}

We also calculated the correlation coefficient between the feature vector composed of each channel value of each pixel coordinate (i, j) and the reference pixel coordinate feature vector. We manually selected two reference pixel locations on the target moving object, and two reference pixel locations on the background. The results are shown in Fig.~\ref{fig:fig8}. The reference pixel coordinates in Fig.~\ref{fig:fig8-a} and Fig.~\ref{fig:fig8-b} are located on the target moving object. It can be seen that the feature map of RiWFPN shows better correlation within the target moving object class than HRFPN. In particular, there is an object that only appears partly on the far right, our method still shows good intra-class correlation of moving objects. Meanwhile, the difference between the target moving object class and the background class in RiWFPN is larger than NASFPN. Therefore, in adverse weather scenarios, RiWNet increases the intra-class correlation of moving objects and minimizes intra-class variance, while enlarging inter-class difference, making the objects easier to segment. 
%这里的图全部重新生成，因为有负值，在*255以及uint8的时候造成了误差，并变成colormap

\subsection{Ablation Studies}
All the results of ablation experiments are performed on VKITTI-moving dataset and obtained by mixing the training sets of three subsets of Fog, Rain, and Illumination together for training, and then evaluating on the testing sets of these three subsets respectively.

\subsubsection{Ablation Studies on RiWFPN}
We verify the effectiveness of each component of RiWFPN, including progressive top-down interaction module (PTI), attention refinement module (ARM) and bottom-up path augmentation (BPA) three components. The results are shown in TABLE~\ref{Tab1}. When there is no attention optimization module (ARM), abundant aggregation information makes the target not significant enough and difficult to discover, so the recall metric is low. The combination of PIL and ARM greatly improves the effect of the network in weather interference. At the same time, BPA also has a positive effect on network accuracy. This ablation study verify our claim of how RiWFPN maintains robustness in weather interference discussed in Section~\ref{sec:RiWFPN}.

\subsubsection{Ablation Studies on RiWNet}
\label{subsubsec:RiWNet}
As discussed in Section~\ref{sec:RiWNet}, in order to reduce the computational burden, we only used the feature maps of 3, 4, and 5 levels into ConvLSTM structure for processing. This ablation study is performed to demonstrate the effectiveness of this design. As shown in TABLE~\ref{Tab2}, using the feature maps of 3, 4, and 5 levels achieves better results than the feature map of only 2 level, while both methods reduce the computational cost.

\subsection{Comparison between RiWFPN and other state-of-the-art FPN structure}
\label{subsubsec:ComparisonFPN}
In order to compare the robustness of RiWFPN and other FPN structures in weather interference, we conduct a series of experiments using RiWNet with different FPN structures. In TABLE~\ref{Tab3}, these are the results obtained by training on the training sets of three subsets of Fog, Rain, and Illumination respectively instead of mixing these three subsets to train. It can be seen that our method obtains the best results in all four metrics in Rain. As a contrast, the results in TABLE~\ref{Tab4} obtained by mixing the training sets of three subsets of Fog, Rain, and Illumination together for training, and then evaluating respectively. It is shown that RiWFPN obtains the best results on all metrics in the three cases of Fog, Rain, and Illumination, which is much better than other FPN methods. This also proves that our method does not need to train only one weather pattern at a time like other methods. It can train the data of multiple weather patterns at one time and still perform better in each weather condition. The qualitative results of RiWNet in VKITTI-moving are shown in Fig.~\ref{fig:fig9}.
\begin{table*}[]	
	\scriptsize	
	\centering	
	\caption{Ablation Studies on RiWFPN. PTI means progressive top-down interaction module, ARM means attention refinement module and BPA means bottom-up path augmentation.}	
	\label{Tab1}
	\begin{tabular}{cccccccccccccccc}
		\toprule
    		\multirow{2}{*}{} & \multicolumn{3}{c}{} & \multicolumn{4}{c}{Fog} 
		& \multicolumn{4}{c}{Rain} & \multicolumn{4}{c}{Illumination} \\
		 \cmidrule(r){5-8} \cmidrule(r){9-12} \cmidrule(r){13-16}
		&  PTI  &  ARM   &   BPA   
		&  R  &  P   &   F   & IoU  
		&  R  &  P   &   F   & IoU  
		&  R  &  P   &   F   & IoU  \\
		\midrule
		RiWNet + RiWFPN    
		&$\surd$ &$-$ &$\surd$  
		&67.8544 &77.1949 &69.2525 &61.0120 
		&67.2736 &77.4963 &68.0752 &61.4972
		&65.1344 &76.9296 &67.3246 &58.8871\\
		\midrule
		RiWNet + RiWFPN    
		&$-$ &$\surd$ &$\surd$  
		&69.0867 &78.2304 &71.3624 &63.0018 
		&69.0227 &77.8654 &71.2289 &63.2614
		&66.6008 &78.1500 &69.6959 &60.9981\\
		\midrule
		RiWNet + RiWFPN    
		&$\surd$ &$\surd$ &$-$  
		&71.7185 &78.0656 &73.4864 &64.9776 
		&71.8159 &78.2153 &73.5866 &65.2354
		&69.3519 &77.6455 &71.6561 &62.8545\\
		\toprule
		RiWNet + RiWFPN    
		&$\surd$ &$\surd$ &$\surd$  
		&\textbf{72.1911} &\textbf{78.7783} &\textbf{73.9263} &\textbf{65.4380} 
		&\textbf{72.3076} &\textbf{78.6134} &\textbf{74.0181} &\textbf{65.7298}
		&\textbf{70.3866} &\textbf{78.5284} &\textbf{72.7195} &\textbf{64.0795}\\
		\bottomrule
	\end{tabular}
	\vspace{0.2cm}
	\footnotesize{}
\end{table*}

\begin{table*}[]	
	\scriptsize	
	\centering	
	\caption{Ablation Studies on RiWNet.}	
	\label{Tab2}
	\begin{tabular}{cccccccccccccc}
		\toprule
    		\multirow{2}{*}{} & \multicolumn{1}{c}{} & \multicolumn{4}{c}{Fog} 
		& \multicolumn{4}{c}{Rain} & \multicolumn{4}{c}{Illumination} \\
		 \cmidrule(r){3-6} \cmidrule(r){7-10} \cmidrule(r){11-14}
		&  Levels used  
		&  R  &  P   &   F   & IoU  
		&  R  &  P   &   F   & IoU  
		&  R  &  P   &   F   & IoU  \\
		\midrule
		RiWNet + RiWFPN    
		&only 2   
		&70.3519 &77.3152 &71.7914 &63.0688
		&70.1986 &77.0337 &71.3784 &63.0978
		&68.5441 &76.8563 &70.4466 &62.0619\\
		\toprule
		RiWNet + RiWFPN    
		&3 4 5  
		&\textbf{72.1911} &\textbf{78.7783} &\textbf{73.9263} &\textbf{65.4380} 
		&\textbf{72.3076} &\textbf{78.6134} &\textbf{74.0181} &\textbf{65.7298}
		&\textbf{70.3866} &\textbf{78.5284} &\textbf{72.7195} &\textbf{64.0795}\\
		\bottomrule
	\end{tabular}
	\vspace{0.2cm}
	\footnotesize{}
\end{table*}

\begin{table*}[]	
	%\scriptsize	
	\centering	
    	\caption{Comparison between RiWFPN and other state-of-the-art FPN structure. These are the results obtained by training on the training sets of three subsets of Fog, Rain, and Illumination respectively and testing on the testing sets of three subsets respectively.}	
	\label{Tab3}
	\begin{tabular}{ccccccccccccc}
		\toprule
		\multirow{2}{*}{} & \multicolumn{4}{c}{Fog} & \multicolumn{4}{c}{Rain} & \multicolumn{4}{c}{Illumination} \\
		\cmidrule(r){2-5} \cmidrule(r){6-9} \cmidrule(r){10-13}
		&  R  &  P   &   F   & IoU
		&  R  &  P   &   F   & IoU  
		&  R  &  P   &   F   & IoU  \\
		\midrule
		RiWNet + FPN    
		&64.3160 &73.2833 &66.7664 &55.9594 
		&63.9901 &75.4874 &67.4494 &57.7026 
		&66.0082 &75.5300 &68.8863 &59.4433\\
		\midrule
		RiWNet + PAFPN  
		&66.5676 &76.0739 &69.3792 &59.4991
		&\textcolor[rgb]{0,0,1}{69.4217} &77.4146 &71.5948 &62.3430 
		&66.9639 &75.7860 &69.4689 &60.1148\\
		\midrule
		RiWNet + FPG    
		&67.2878 &75.9978 &70.0873 &60.9971
		&69.1623 &78.0225 &\textcolor[rgb]{0,0,1}{72.1775} &\textcolor[rgb]{0,0,1}{62.7113} 
		&64.4351 &75.0395 &66.6059 &57.6032\\
		\midrule
		RiWNet + RFP    
		&62.7925 &73.0408 &66.0832 &55.4460 
		&67.2318 &76.6518 &69.4800 &59.8921 
		&\textcolor[rgb]{1,0,0}{68.6223} &76.8415 &\textcolor[rgb]{1,0,0}{71.0885} &\textcolor[rgb]{1,0,0}{61.8353}\\
		\midrule
		RiWNet + NASFPN 
		&\textcolor[rgb]{0,0,1}{71.0906} &\textcolor[rgb]{1,0,0}{79.9115} &\textcolor[rgb]{1,0,0}{73.8307} &\textcolor[rgb]{0,0,1}{63.4506} 
		&65.0104 &74.0085 &67.1104 &57.1253 
		&66.1716 &\textcolor[rgb]{1,0,0}{81.4152} &\textcolor[rgb]{0,0,1}{70.6771} &59.8121\\
		\midrule
		RiWNet + HRFPN  
		&\textcolor[rgb]{1,0,0}{71.1792} &\textcolor[rgb]{0,0,1}{78.8042} &\textcolor[rgb]{1,0,0}{73.6642} &\textcolor[rgb]{1,0,0}{64.7437} 
		&68.7144 &\textcolor[rgb]{0,0,1}{78.2312} &71.7578 &62.3309 
		&67.3544 &76.5178 &69.8822 &60.5343\\
		\toprule
		RiWNet + RiWFPN  
		&68.7513 &76.6170 &70.9883 &61.6493 
		&\textcolor[rgb]{1,0,0}{70.3652} &\textcolor[rgb]{1,0,0}{79.6420} &\textcolor[rgb]{1,0,0}{73.3892} &\textcolor[rgb]{1,0,0}{64.5407} 
		&\textcolor[rgb]{0,0,1}{68.0239} &\textcolor[rgb]{0,0,1}{77.0120} &70.2856 &\textcolor[rgb]{0,0,1}{61.2063}\\
		\bottomrule
	\end{tabular}
	\vspace{0.2cm}
	\footnotesize{
		\\ Best results are highlighted in $\rm \textcolor[rgb]{1,0,0}{red}$ with second best in $\rm \textcolor[rgb]{0,0,1}{blue}$.}
\end{table*}

\begin{table*}[]	
	%\scriptsize	
	\centering	
	\caption{Comparison between RiWFPN and other state-of-the-art FPN structure. These results are obtained by mixing the training sets of three subsets of Fog, Rain, and Illumination together for training, and then evaluating respectively.}	
	\label{Tab4}
	\begin{tabular}{ccccccccccccc}
		\toprule
		\multirow{2}{*}{} & \multicolumn{4}{c}{Fog} & \multicolumn{4}{c}{Rain} & \multicolumn{4}{c}{Illumination} \\
		\cmidrule(r){2-5} \cmidrule(r){6-9} \cmidrule(r){10-13}
		&  R  &  P   &   F   & IoU
		&  R  &  P   &   F   & IoU  
		&  R  &  P   &   F   & IoU  \\
		\midrule
		RiWNet + FPN    
		&67.1934 &75.9910 &69.6194 &60.2508 
		&68.0027 &76.0370 &70.2864 &60.8403 
		&64.9424 &75.2914 &67.6752 &57.9608\\
		\midrule
		RiWNet + PAFPN  
		&69.8223 &76.8181 &71.9257 &63.1057
		&68.7693 &76.0352 &70.9302 &62.0728 
		&67.0133 &75.7977 &69.3109 &60.0565\\
		\midrule
		RiWNet + FPG    
		&66.0783 &76.4919 &68.4716 &59.5495
		&65.5916 &76.6606 &68.3113 &59.3429 
		&64.4074 &76.6008 &67.3842 &58.0600\\
		\midrule
		RiWNet + RFP    
		&69.6138 &76.6718 &71.8139 &62.5276 
		&68.9629 &75.2874 &70.6670 &61.2696 
		&\textcolor[rgb]{0,0,1}{67.6848} &75.2554 &\textcolor[rgb]{0,0,1}{69.7875} &\textcolor[rgb]{0,0,1}{60.4021}\\
		\midrule
		RiWNet + NASFPN 
		&69.3932 &\textcolor[rgb]{0,0,1}{77.6686} &71.4881 &62.7676
		&68.8042 &\textcolor[rgb]{0,0,1}{78.0920} &71.0696 &62.1259 
		&64.0359 &\textcolor[rgb]{0,0,1}{77.1758} &67.1837 &57.5821\\
		\midrule
		RiWNet + HRFPN  
		&\textcolor[rgb]{0,0,1}{70.9928} &77.4495 &\textcolor[rgb]{0,0,1}{72.9079} &\textcolor[rgb]{0,0,1}{63.8511}
		&\textcolor[rgb]{0,0,1}{70.4500} &76.8053 &\textcolor[rgb]{0,0,1}{72.2372} &\textcolor[rgb]{0,0,1}{63.2345}
		&67.2949 &75.7817 &69.6622 &60.2593\\
		\toprule
		RiWNet + RiWFPN  
		&\textcolor[rgb]{1,0,0}{72.1911} &\textcolor[rgb]{1,0,0}{78.7783} &\textcolor[rgb]{1,0,0}{73.9263} &\textcolor[rgb]{1,0,0}{65.4380} 
		&\textcolor[rgb]{1,0,0}{72.3076} &\textcolor[rgb]{1,0,0}{78.6134} &\textcolor[rgb]{1,0,0}{74.0181} &\textcolor[rgb]{1,0,0}{65.7298}
		&\textcolor[rgb]{1,0,0}{70.3866} &\textcolor[rgb]{1,0,0}{78.5284} &\textcolor[rgb]{1,0,0}{72.7195} &\textcolor[rgb]{1,0,0}{64.0795}\\
		\bottomrule
	\end{tabular}
	\vspace{0.2cm}
	\footnotesize{
		\\ Best results are highlighted in $\rm \textcolor[rgb]{1,0,0}{red}$ with second best in $\rm \textcolor[rgb]{0,0,1}{blue}$.}
\end{table*}

\subsection{Comparison with Prior Works}
\subsubsection{FBMS}
\begin{figure*}
	\begin{center}
		\includegraphics[width=1\linewidth]{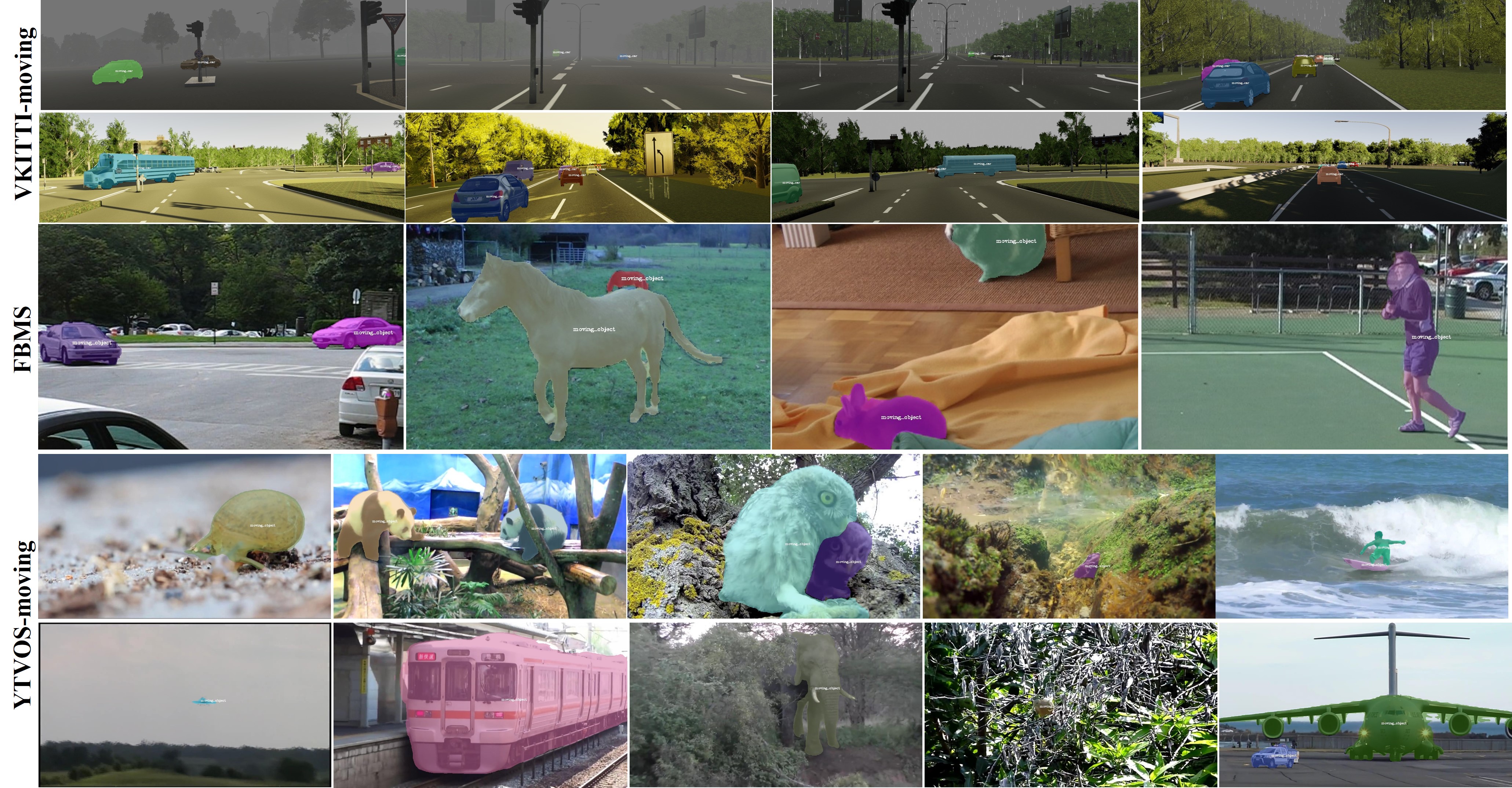}
	\end{center}
	\caption{Qualitative results for three datasets.}
	\label{fig:fig9}
\end{figure*}

FBMS is a widely used moving object segmentation dataset, and RiWNet is evaluated on this dataset against multiple prior works. RiWNet is evaluated on testing set of the standard FBMS using the model trained from training set mixed by FBMS and YTVOS. The results are shown in TABLE~\ref{Tab05}. RiWNet performs the best in precision and F-measure and outperforms other methods by over 3.7\% and 1.6\%, respectively. In terms of recall, it also outperforms most methods and outperforms other methods except U$^2$-ONet~\cite{U2ONet2021} by over 0.9\%. The qualitative results are shown in Fig.~\ref{fig:fig9}.

\begin{table}[h]	
	\centering	
	\caption{FBMS results using the official metric}	
	\label{Tab05}
	\begin{tabular}{ccccc}
		\toprule
		\multirow{1}{*}{} & \multicolumn{4}{c}{Multi-object Motion Segmentation} \\
		\cmidrule(r){2-5} 
		&  R      &  P   &   F & IoU \\
		\midrule
		CCG~\cite{bideau2018best}            &63.07   &74.23   &64.97  &-- \\
		STB~\cite{shen2018submodular}        &66.53   &87.11   &75.44  &-- \\
		OBV~\cite{xie2019object}             &66.60   &75.90   &67.30   &--\\
		TSA~\cite{dave2019towards}           &80.40   &\textcolor[rgb]{0,0,1}{88.60}   &\textcolor[rgb]{0,0,1}{84.30} &-- \\
		U$^2$-ONet~\cite{U2ONet2021}         &\textcolor[rgb]{1,0,0}{83.10}   &84.80   &81.84   &\textcolor[rgb]{1,0,0}{79.70}  \\
		RiWNet      &\textcolor[rgb]{0,0,1}{81.36}    &\textcolor[rgb]{1,0,0}{92.30}   &\textcolor[rgb]{1,0,0}{85.99}   &\textcolor[rgb]{0,0,1}{76.71} \\
		\bottomrule
	\end{tabular}
	\vspace{0.2cm}
	\footnotesize{
		\\ Best results are highlighted in $\rm \textcolor[rgb]{1,0,0}{red}$ with second best in $\rm \textcolor[rgb]{0,0,1}{blue}$.}
\end{table}

\subsubsection{YTVOS-moving}
RiWNet is further evaluated on the YTVOS-moving testing set using the model trained from the YTVOS-moving training set, as defined in~\cite{dave2019towards}. The results are listed in TABLE~\ref{Tab06}. RiWNet performs the best in recall, precision and F-measure, and outperforms U$^2$-ONet~\cite{U2ONet2021} and TSA~\cite{dave2019towards} by over 4.1\% in precision and over 3.4\% in F-measure. The qualitative results are shown in Fig.~\ref{fig:fig9}.
\begin{table}[h]	
	\scriptsize	
	\centering	
	\caption{Results for the YouTube Video Object Segmentation (YTVOS)-Moving dataset.}	
	\label{Tab06}
	\begin{tabular}{ccccc}
		\toprule
		\multirow{1}{*}{} & \multicolumn{4}{c}{Multi-Object Motion Segmentationn} \\
		\cmidrule(r){2-5} 
		&  R    &  P   &  F & IoU \\
		\midrule
		TSA~\cite{dave2019towards}      &66.40   &74.50    &68.30   &$–$    \\
		U$^2$-ONet~\cite{U2ONet2021}    &70.56   &74.64    &69.93   &\textbf{65.67}  \\
		RiWNet                          &\textbf{70.73}   &\textbf{79.80}    &\textbf{73.35}   &65.02  \\
		\bottomrule
	\end{tabular}
\end{table}

\subsection{Applications}
Main applications of RiWNet are dynamic Visual SLAM or visual-LiDAR fusion odometry/SLAM as well as 3D dense mapping. Here we show the effectiveness of RiWNet, adding RiWNet as a processing module to segment moving objects for keyframes in our previous work called DV-LOAM~\cite{DV-LOAM2021} of visual-LiDAR fusion SLAM. In DV-LOAM, because the relative transformation of the camera and laser is known, just using the image of the moving object segmentation result can handle the point cloud of the entire Visual LiDAR fusion SLAM. We only use masks to remove effectively all visual feature points and point clouds belonging to moving objects and no further operations have been employed. We conduct experiments on the Sequence 04 in KITTI odometry benchmark~\cite{KITTI2012} because this sequence contains more moving cars. Because RiWNet runs in another thread and only processes key frames, the experimental results show that the adding of RiWNet (running about 5HZ) does not affect the real-time operation of DV-LOAM.

\subsubsection{visual-LiDAR fusion odometry/SLAM}
We compare the improved DV-LOAM after adding RiWNet to standard DV-LOAM in TABLE~\ref{Tab07}. The results show that estimating odometry using the improved DV-LOAM are higher precision than standard DV-LOAM. 

\begin{table}[h]	
	\scriptsize	
	\centering	
	\caption{Comparison of odometry results.}	
	\label{Tab07}
	\begin{tabular}{ccc}
		\toprule
		\multirow{1}{*}{} & \multicolumn{2}{c}{Approach} \\
		\cmidrule(r){2-3} 
		&  DV-LOAM   &  DV-LOAM+RiWNet \\
		\midrule
		Sequence 04      &0.30/0.61   &\textbf{0.26/0.56}    \\
		\bottomrule
	\end{tabular}
	\vspace{0.2cm}
	\footnotesize{
		\\ Relative errors averaged over trajectories in Sequence 04: relative
		rotational error in degrees per 100 m / relative translational error in \%.}
\end{table}

\subsubsection{3D Mapping}
\begin{figure*}
	\subfigure[] {
		\label{fig:fig10-a}     
		\includegraphics[width=0.48\linewidth]{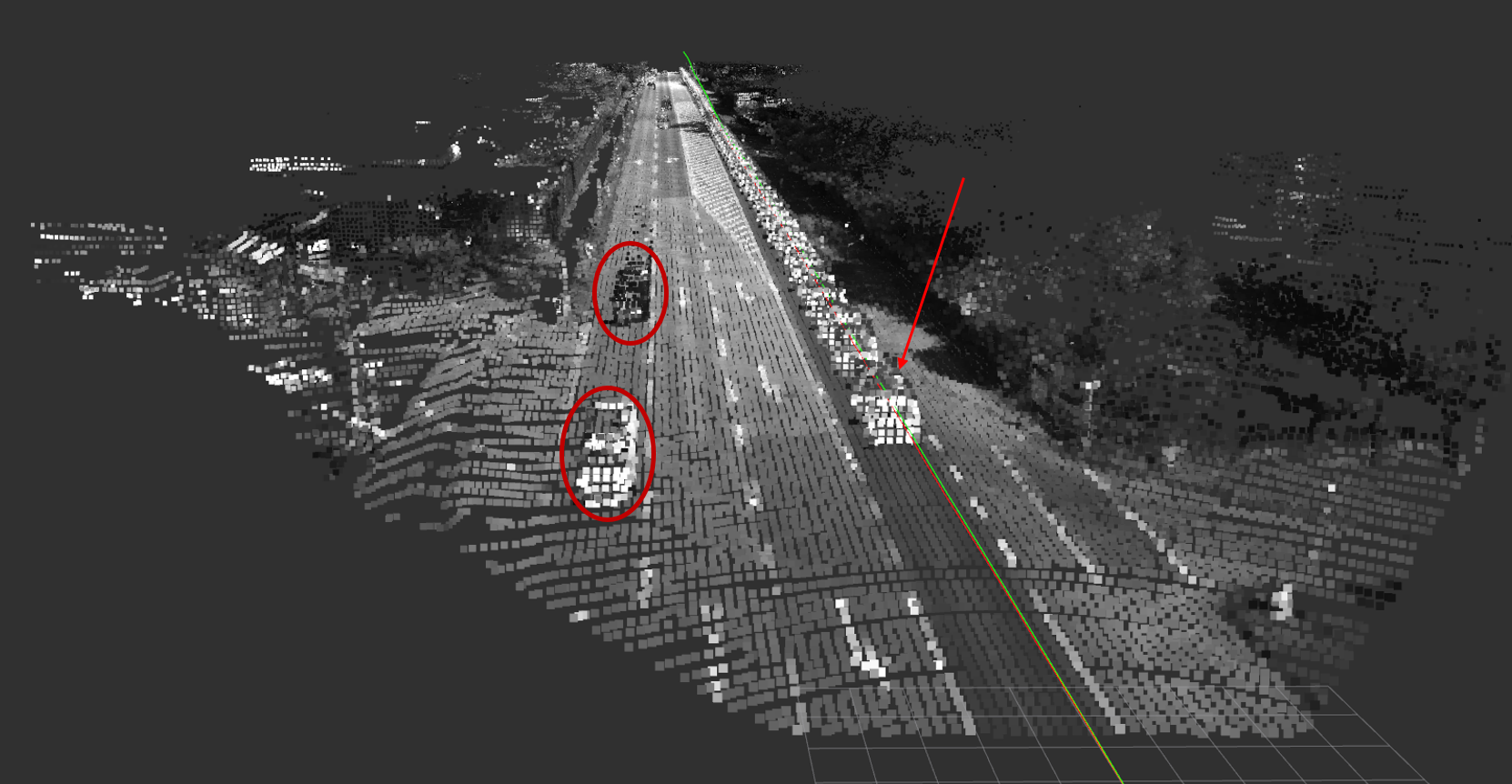}  
	} 
%	\hspace{-0.2cm}
	\subfigure[] {
		\label{fig:fig10-b}     
		\includegraphics[width=0.48\linewidth]{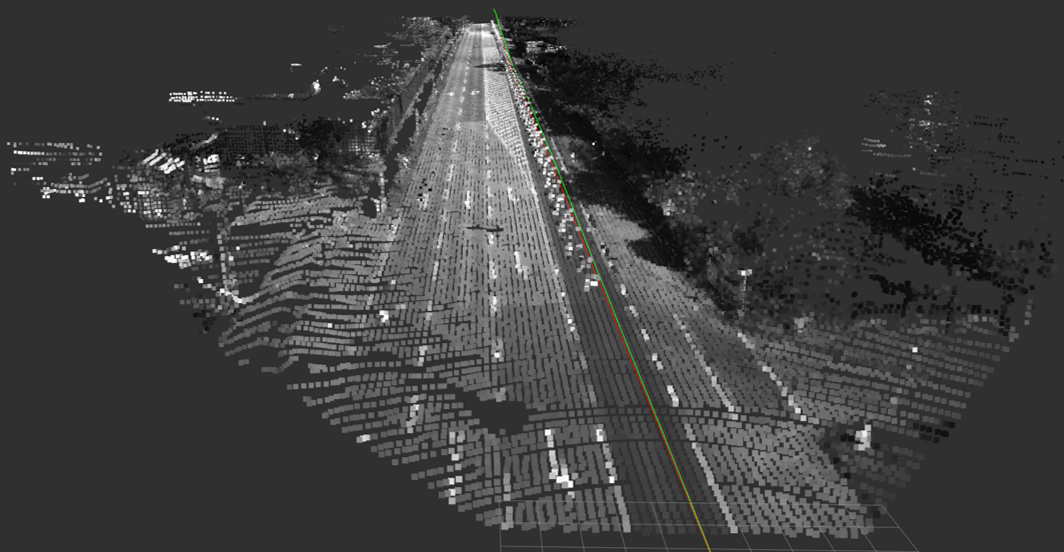}  
	} 
	\vspace{-0.2cm}
	\caption{Generated point cloud maps on Sequence 04. (a) shows the result without using moving object masks. (b) shows the result with using moving object masks for removing point cloud predicted as moving. In (a), the red ellipse and red arrow in the figure indicate the moving objects included in the final map. In (b), the point cloud of moving objects is almost removed. There is a certain point cloud of moving objects at the place pointed by the red arrow, because there is a certain region at the edge of the object that is difficult to segment.}
	\label{fig:fig10}
	\vspace{-0.5cm}
\end{figure*}

In Fig.~\ref{fig:fig10}, We show the final point cloud maps generated with and without using moving object masks respectively. As shown in Fig.~\ref{fig:fig10-a}, a polluted point cloud map containing moving objects is generated due to the existence of moving objects. This polluted map is likely to reduce the accuracy of loop closure detection and the effect of motion planning. By using our RiWNet to handle moving objects, the static structure are more clearly observable. 

\section{Conclusion}
The effect of moving object segmentation usually decrease enormously in adverse weather conditions as compared to good weather conditions. In this paper, we firstly propose a novel RiWFPN combining a progressive top-down interaction and attention refinement module to strengthen the low-frequency structure information of the network. Compared with other FPN structures, using RiWFPN as a neck structure can improve the robustness of the network in degrading weather conditions. We then extend SOLOV2 to learn temporal motion information and propose a novel end-to-end moving object instance segmentation network, called RiWNet. RiWNet uses a pair of adjacent RGB frames as inputs and performs robustly in different weather environments by integrating RiWFPN. We construct a moving object instance segmentation dataset considering different weather conditions for verifying the effectiveness of the method. Experimental results fully demonstrate how RiWNet enhance the feature map to improve the robustness of the network in the interference of weather conditions. We also show that RiWNet achieves state-of-the-art performance in some challenging datasets, especially in harsh weather scenarios. 

In the near future, we consider incorporating RiWNet to our previous work: a stereo dynamic visual SLAM system~\cite{DymSLAM2021}, as the module of moving object segmentation. We will use RiWNet to segment moving objects in key-frames of SLAM and propose a real-time end-to-end dynamic SLAM be capable of estimating simultaneously global trajectories of the camera and moving objects.

% or


\ifCLASSOPTIONcaptionsoff
  \newpage
\fi



\bibliographystyle{IEEEtran}
\bibliography{IEEEabrv,egbib}



%\newpage


%\vfill


\end{document}